# Random Silicon Sampling: Simulating Human Sub-Population Opinion Using a Large Language Model Based on Group-Level Demographic Information

Seungjong Sun[1], Eungu Lee[1], Dongyan Nan[2], Xiangying Zhao[2], Wonbyung Lee[1],
Bernard J. Jansen[3], Jang Hyun Kim[1,2]

[1]Department of Applied Artificial Intelligence, [2]Department of Interaction Science, Sungkyunkwan University
[3]Qatar Computing Research Institute, Hamad Bin Khalifa University

{tmdwhd406, dldmsrn0516, zxy94, co2797}@g.skku.edu, {ndyzxy0926, alohakim}@skku.edu,
jjansen@acm.org

## Abstract

Large language models exhibit societal biases associated with demographic information, including race, gender, and others. Endowing such language models with personalities based on demographic data can enable generating opinions that align with those of humans. Building on this idea, we propose "random silicon sampling," a method to emulate the opinions of the human population sub-group. Our study analyzed 1) a language model that generates the survey responses that correspond with a human group based solely on its demographic distribution and 2) the applicability of our methodology across various demographic subgroups and thematic questions. Through random silicon sampling and using only group-level demographic information, we discovered that language models can generate response distributions that are remarkably similar to the actual U.S. public opinion polls. Moreover, we found that the replicability of language models varies depending on the demographic group and topic of the question, and this can be attributed to inherent societal biases in the models. Our findings demonstrate the feasibility of mirroring a group's opinion using only demographic distribution and elucidate the effect of social biases in language models on such simulations.

## 1 Introduction

With the advent of large language models (LLMs), the societal biases inherent in these models have been extensively researched (Kolisko and Anderson, 2023; Zhou et al., 2022; Feng et al., 2023; Santy et al., 2023). These models learn human-like biases associated with race, gender, ethnicity, and others from human-written data (Schramowski et al., 2023; Peters and Matz, 2023). Although many studies have attempted to mitigate societal biases in LLMs (Barocas and Selbst, 2016; Mayson, 2018; Panch, Mattie, and Atun, 2019), recent research has explored methods to retrieve and induce these biases for understanding human opinions (Santurkar et al., 2023; Chu et al., 2023; Mills et al., 2023). Opinions about or from the perspective of certain human groups can be generated by models by conditioning the inherent societal biases of the model (Jiang et al., 2022; Simmons, 2022; Caron and Srivastava 2022; Hartmann et al., 2023).

These attempts to leverage the biases of LLMs have attracted the attention of social science researchers who conduct public opinion polls and surveys. The surveys commonly used in social science research were constrained in terms of the high cost and reliability of responses (Shapiro, 2019; Keeter et al., 2017; Kennedy et al., 2018). Consequently, several studies attempted to overcome the limitations of traditional surveys that use LLMs (Jansen et al., 2023; Brand et al., 2023; Grossmann et al., 2023). For example, some studies focused on conditioning language models with demographic information of human sub-populations for investigating whether language models can generate responses similar to actual respondents (Hwang et al., 2023; Kalinin, 2023) and simulate group-level opinions (Santurkar et al., 2023; Dominguez-Olmedo et al., 2023; Park et al., 2023; Bisbee et al., 2023).

In line with these studies, Argyle et al. (2023) introduce the "silicon sampling" methodology, which enables the generation of survey responses similar to those of humans using language models, and they discuss the potential of language models as surrogates for survey respondents. Silicon sampling has limitations because it conditions LLMs with individual-level demographic information, which cannot be determined before conducting a survey. Our study proposes "random silicon sampling," which generates the opinions correspond with a real human group through LLMs based on the demographic distribution of the group without requiring individual-level data to address this research gap.

In our experiments, we created a "random silicon sample," which is a group of randomly generated synthetic respondents that follow the demographic distribution of a human sub-population. We obtained survey responses by prompting the model with the demographic information of synthetic individual respondents, along with survey questions. Through our experiment, we examined how well the random silicon sample generated the response distribution aligned with the real human group's opinion.

Furthermore, we conducted experiments in various settings to explore the generalizability of our methodology. We first investigated whether response generation using a language model varied among specific subgroups within the population. Further, we explored the sample size required to simulate the opinions of the target group effectively. Finally, we examined whether random silicon sampling could generate opinions across various topics.

We proposed the possibility of using a language model to generate opinions that accurately mirror specific groups based on their demographic distribution. We confirmed that the replicability of such language models varies depending on the target group and question, which revealed that these replication patterns are affected by inherent societal biases in the language models.

The contributions of this research are as follows:
1) Introduced a novel methodology for survey augmentation using LLMs that reduces the cost and time of conducting surveys and pretests.
2) Demonstrated the feasibility of predicting a group's opinion using only demographic information, without requiring individual-level data.
3) Clarified biases inherent in language models towards specific demographic groups and survey topics that must be considered for the generalized application of this approach, thereby providing insights into the generalizability and limitations of the method.

## 2 Related Work

**Simulate human opinion** Existing studies discovered that LLMs exhibit biases regarding genders, races, nationalities, and in social contexts (Feng et al., 2023; Durmus et al., 2023; Santurkar et al., 2023). In addition, studies leverage biases in these language models for simulating actual human opinions, which implies that it is possible to mirror a real human's opinion by conditioning a model to "virtual individuals" with specific identity and personality profiles (Beck et al., 2023; Deshpande et al., 2023; Caron and Srivastava 2022; Jinag et al., 2023). Existing research explores the potential of prompting LLMs with personalities to mirror human opinions, behaviors, and attitudes (Cheng et al., 2023b; Gupta et al., 2023; Salewski et al., 2023; Jiang et al. 2022; Hwang et al., 2023; Kalinin, 2023), especially in politics (Törnberg et al., 2023; Wu et al., 2023; Perez et al. 2022; Simmons 2022), economics (Motoki et al., 2023; Horton, 2023; Brand et al., 2023), and psychology (Pellert et al., 2023; Griffin et al., 2023; Hagendorff et al., 2022).

Building on the potential of language models to align with human opinions, research has been conducted on using LLMs to augment surveys. The use of LLMs has been explored for effectively simulating surveys, tasks, and experiments in traditional social science research such as behavioral science and psychology (Aher et al., 2023; Dillion et al., 2023; Binz and Schulz, 2023; Rosenbusch et al., 2023; Bail, 2023). Further studies explored the possibility of replacing survey respondents with language models conditioned to the personalities of the respondents. Kim and Lee (2023) demonstrated the feasibility of fine-tuning LLMs with the personal information of individual and the previous response patterns for missing data imputation and retrodiction in public opinion poll data. Santurkar et al. (2023) investigated if LLMs prompted with information about a sub-population can represent the opinions of that group. Durmus et al. (2023) found that LLMs prompted to answer from the perspective of a specific country generate opinions that align more closely with the prompted population on societal issues. Meanwhile, research focused on the limitations of LLMs as surrogates for survey respondents, including bias towards labeling and orders of answer choices (Dominguez-Olmedo et al., 2023; Kalinin, 2023), extremity of responses (Bisbee et al., 2023; Park et al., 2023), and bias toward specific sub-populations (Santurkar et al., 2023). Our study explores the feasibility and limitations of LLMs as substitutes for survey respondents, investigating deployability in social science research, across various demographic sub-groups and topics.

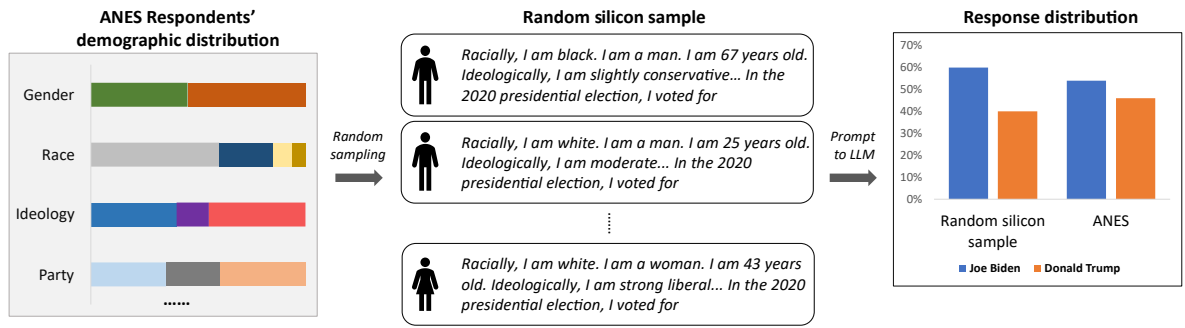

Figure 1: Pipeline of random silicon sampling: 1) Extract the demographic distribution of the target sub-group. 2) Extract demographic information randomly from the demographic distributions of the target group to configure a random silicon subject. 3) Input demographic information of the random silicon subjects as a prompt into the language model with a survey question. 4) Obtain distributions of responses conditioned with the demographic information of random silicon subjects.

**Silicon sampling** Argyle et al. (2023) proposed the method of "silicon sampling" based on the capability of LLMs to predict human opinions using demographic information; this method employs LLMs to forecast the opinions of specific sub-populations. Argyle et al. (2023) configured the model to create a "silicon subject" for each human participant and had these artificial subjects undertake the same tasks presented to them. This study aimed to investigate if the language model can simulate opinions at the group level when the data size is adequately large, while the generated responses at the individual level may not necessarily align closely with the actual responses. Moreover, their experimental results indicated that the silicon samples produced responses highly correlated (>0.90) with actual human responses when modeled after respondents from the American National Election Studies (ANES) with matching demographic data. Subsequent studies investigated the potential for expanding silicon sampling (Sanders et al., 2023; Lee et al., 2023).

However, this silicon sampling requires demographic information about individuals within the group, which is challenging to obtain in actual survey scenarios. In the survey-based research, the researchers conducted surveys on samples with demographic distributions similar to those of the target population for investigating the opinions of the population (Miller and Brewer, 2003; Cobanoglu et al., 2021; Ortega 2015). Therefore, this study introduces a "random silicon sampling" approach that generates opinions closely aligned with human sub-populations. Random silicon sampling uses LLMs to simulate the opinions of human subgroups based on their demographic distributions.

## 3 Methods

The proposed random silicon sampling method simulates the response distribution of a specific human subpopulation by conditioning LLMs on the demographic distribution of the group. Figure 1 illustrates the pipeline of our experiments. We expanded the silicon sampling experimental setup described by Argyle et al. (2023). We also conducted experiments on stratified sampling, down-sampling, and multiple-question cases to discuss the generalizability of random silicon sampling. The dataset and code to use random silicon sampling and reproduce our results is at https://anonymous.4open.science/r/RSS-1D4D.

### 3.1 Data

We utilize the ANES dataset, which is the primary source for comprehending the American public opinion (ANES 2021; Argyle et al., 2023). We leverage the ANES 2020 pre-election sample consisting of interviews with respondents between August 18, 2020, and November 3, 2020. The sample includes demographic information, responses to presidential votes, and political surveys from 5,441 respondents. To generate the response distribution correlate with ANES respondents, we extract the distribution of the demographic variables of each respondent, including race/ethnicity, gender, age, ideological self-placement, party identification, political interest, church attendance, and discussing politics with family and friends (Argyle et al., 2023).

## 3.2 Random silicon sampling

We configure random silicon subjects with randomly selected demographic information based on the demographic distribution of the ANES respondents to condition a language model for generating a response distribution. Considering that $A_{ANES}$ represents the actual response distribution of ANES respondents to survey questions, our experiment conditions a language model using the demographic distribution of ANES respondents, $D_{ANES}$, to generate $A_{RSS}$. We compute the demographic distributions for each of the eight demographics variable $\{D_{race}, D_{gender}, D_{age}, ..., D_{discussing\ politics}\} = D_{ANES}$, and subsequently, we randomly extracted demographic information to create random silicon subjects represented as $R_i = \{d_{race\ i} \sim D_{race}, d_{gender\ i} \sim D_{gender}, d_{age\ i} \sim D_{age}, ..., d_{discussing\ politics\ i} \sim D_{discussing\ politics}\}$.
The demographic details of $R_i$ are portrayed in the first person and fed into the model as prompts along with the survey questions. Finally, the language model yields a survey response. We reiterate this method to obtain $A_{RSS}$, which represents the distribution of responses from random silicon subjects.

## 3.3 Evaluation

We compute the similarity between the two response distributions, $A_{ANES}$ and $A_{RSS}$ to assess the replicability of the language model to simulate the response distribution of actual ANES respondents.

**Chi-Square Test for Homogeneity** This test was applied to ascertain the presence of statistically significant differences between response distributions. P<0.05 indicates that the two response distributions are statistically significantly different.

**Kullback-Leibler Divergence** (KL-divergence) We calculated the KL-divergence to measure the similarity between response distributions. The KL-divergence values span [0, ∞], where the lower values indicate a closer resemblance between the response distribution of the random silicon sample and the actual ANES response distribution (Leemann et al., 2021).

| Sample | Biden rate | Trump rate | Chi-squared value | KL-divergence |
|---|---|---|---|---|
| ANES 2020 | 58.88 % | 41.18 % | | |
| Silicon sample | 55.61 % | 44.39 % | 8.8931 * | 0.00210 |
| RSS 1 | 58.00 % | 42.00 % | 0.5688 | 0.00014 |
| RSS 2 | 57.99 % | 42.01 % | 0.5897 | 0.00014 |
| RSS 3 | 57.85 % | 42.15 % | 0.8107 | 0.00020 |
| RSS 4 | 57.73 % | 42.27 % | 1.0182 | 0.00024 |
| RSS 5 | 57.45 % | 42.55 % | 1.5668 | 0.00039 |
| RSS 6 | 57.27 % | 42.73 % | 2.0724 | 0.00049 |
| RSS 7 | 57.24 % | 42.76 % | 2.1671 | 0.00051 |
| RSS 8 | 57.19 % | 42.81 % | 2.3121 | 0.00055 |
| RSS 9 | 57.02 % | 42.98 % | 2.8054 | 0.00066 |
| RSS 10 | 56.61 % | 43.39 % | 4.2774 * | 0.00100 |

Table 1: Replicability of random silicon sampling. Chi-squared test p-values <0.05 with * indicate that there is a statistically significant difference between the response distributions of the sample and ANES. The closer the KL-divergence is to zero, the more similar are the two response distributions.

## 4 Experimental Settings

Our experiments were conducted using the GPT-3.5-turbo API. The demographic variables for each random silicon subject were described in the first person and entered as `system prompts`, whereas the survey questions were provided as `user prompts` in the API calling. Table A2 lists the examples of prompts. Finally, the model generated responses to the survey questions. We generate responses equivalent to the number of respondents in the ANES 2020 (=5441).

### 4.1 Response generating process

We generated responses regarding voting choices for the 2020 U.S. presidential election using the demographic distribution of ANES 2020 respondents to compare random silicon sampling with silicon sampling. The `MAX TOKEN` for generation was set to two for limiting responses to the names of the candidates. Among the generated responses, phrases such as [*Joe Biden, Joe, Biden, the Democratic, a Democratic*] were coded as a vote for "Joe Biden," and [*Donald Trump, Donald, Trump, the Republican, a Republican*] were coded as a vote for "Donald Trump." We conducted additional experiments that included changing the order of the prompts (Kalinin, 2023) using ANES data from different years and testing across different models to verify the robustness of the methodology.

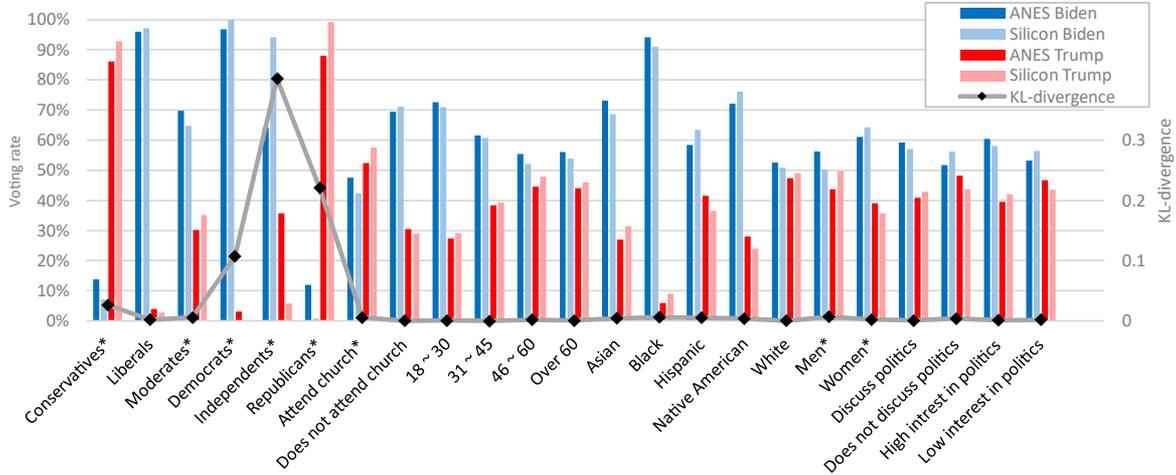

Figure 2: Results of stratified sampling experiments for each demographic subgroup. Dark blue and red colors indicate voting choices of ANES respondents, while light blue and red colors represent voting choices generated by random silicon sample. Chi-Square Test p-value <0.05 with * indicates that there is statistically significant difference between response distribution of sample and ANES response distribution. Black diamond-shaped dots signify KL-divergence values for each group.

### 4.2 Further analysis

**Stratified sampling** We classified eight demographic variables into 23 subgroups (e.g., Men, Women, White, Black, …) and conducted experiments accordingly to compare the opinion replicability of random silicon sampling for specific groups. We calculated the demographic distribution of subgroups filtered by each variable in ANES data and performed random silicon sampling based on these distributions

**Down sampling** We conducted down sampling experiments to determine the appropriate sample size for random silicon sampling. Based on the original sample size of 5,441 samples, we performed experiments with sample sizes decreasing from 90 % to 10 % and from 9 % to 1 % by incorporating the demographic distribution of ANES 2020 respondents.

**Multiple questions** We tested whether random silicon sampling could be applied to issues beyond presidential candidate selection. From the surveys included in ANES 2020, we selected 10 multiple-choice questions on various topics referring to previous studies (Hwang et al., 2023; Kalinin, 2023;

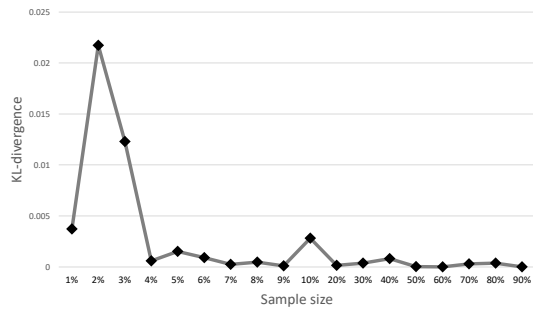

Figure 3: Results of Down Sampling experiments.

Feng et al., 2023). MAX TOKEN was set to one to restrict responses to the given options. Detailed descriptions for questions and prompts used in the experiment can be found in Appendix A3.

## 5 Results

We evaluate whether response distribution generated through random silicon sampling based on the demographic distribution of ANES respondents, corresponds to their actual response distribution of ANES respondents. Further, we verified the consistency of the results and confirmed their applicability across various human subgroups and survey topics on different subjects.

Table 1 illustrates the voting choices for Joe Biden and Donald Trump in the 2020 Presidential Election among the actual ANES respondents compared to the choices derived from silicon sampling and random silicon samplings. Table 1 illustrates that, through random silicon sampling, the language model generated a response distribution that is remarkably similar to the actual ANES responses (average KL-divergence = 0.0004), exhibiting higher similarity than the silicon sample. Further, the voting choice distributions from ten repeated random silicon samplings in the same setting demonstrated high consistency with an average vote rate for Biden of 57.43% and a standard deviation of 0.46%. Although one out of ten repetitions exhibited a p-value below 0.05, all ten exhibited very low KL-divergences. This confirms that random silicon sampling can produce response distributions that closely resemble actual ANES respondents and that the results are reproducible. The experimental results for different settings of prompts, ANES data years, and models can be found in Appendix B.

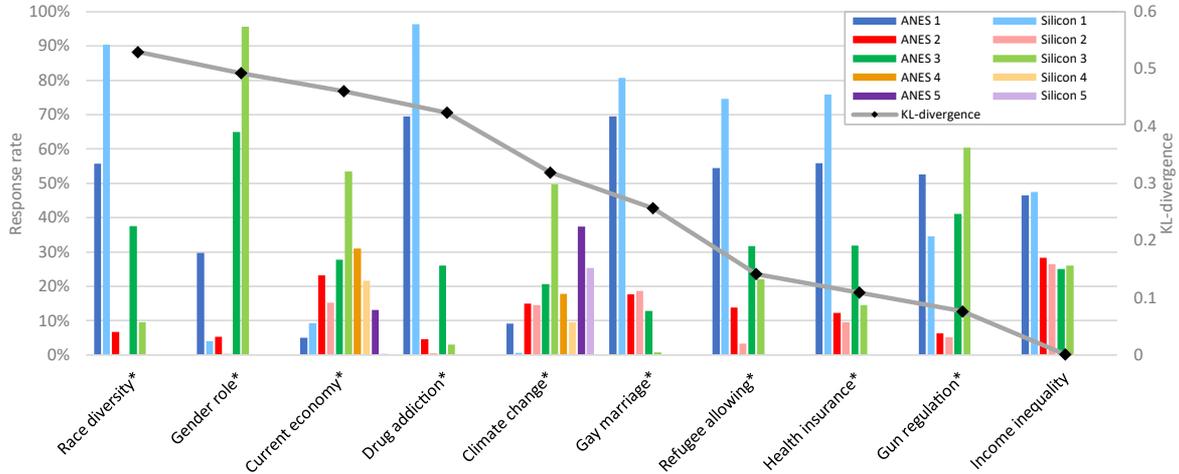

Figure 4: Results of multiple-choice question experiments on 10 different topics. Darker colors represent choices of ANES respondents, while lighter colors indicate choices generated by random silicon sample. Chi-squared test p-values <0.05 with * indicate that there are statistically significant differences between the response distributions of the sample and ANES. The black diamond-shaped dots represent the KL-divergence values for each group.

| Groups | Gender role | Current economy | Race diversity | Drug addiction | Climate change | Gay marriage | Refugee allowing | Gun regulation | Health insurance | Income inequality | AVG |
|---|---|---|---|---|---|---|---|---|---|---|---|
| White | 0.3521 | 0.3702 | 0.4492 | 0.4341 | 0.3658 | 0.2101 | 0.1061 | 0.0874 | 0.0905 | 0.0000 | **0.2465** |
| Republican | 0.1894 | 0.2244 | 0.4153 | 0.4745 | 0.4598 | 0.2884 | 0.1278 | 0.1609 | 0.1457 | 0.1083 | **0.2594** |
| Man | 0.4956 | 0.4554 | 0.3100 | 0.3864 | 0.4052 | 0.2384 | 0.1532 | 0.0788 | 0.0799 | 0.0039 | **0.2607** |
| Over 60 | 0.3213 | 0.2792 | 0.5853 | 0.5019 | 0.3591 | 0.1659 | 0.1711 | 0.0467 | 0.1754 | 0.0114 | **0.2617** |
| Conservatives | 0.1616 | 0.1992 | 0.4119 | 0.6618 | 0.4438 | 0.2727 | 0.1398 | 0.1214 | 0.1991 | 0.0935 | **0.2705** |
| Does not attend | 0.6150 | 0.4576 | 0.4946 | 0.4376 | 0.2890 | 0.1689 | 0.1544 | 0.0502 | 0.1048 | 0.0083 | **0.2780** |
| No interest | 0.3745 | 0.6217 | 0.3327 | 0.0872 | 0.6046 | 0.3798 | 0.0541 | 0.2737 | 0.0166 | 0.0462 | **0.2791** |
| 18~30 | 0.7642 | 0.4324 | 0.4789 | 0.2316 | 0.2829 | 0.3103 | 0.1090 | 0.1411 | 0.0663 | 0.0138 | **0.2831** |
| Like to discuss | 0.5271 | 0.4603 | 0.4374 | 0.5070 | 0.3429 | 0.2405 | 0.1596 | 0.0573 | 0.1204 | 0.0026 | **0.2855** |
| Native American | 0.7184 | 0.3407 | 0.2546 | 0.5551 | 0.2231 | 0.1776 | 0.2106 | 0.2165 | 0.1795 | 0.0112 | **0.2887** |
| Attend church | 0.2859 | 0.4206 | 0.6053 | 0.5934 | 0.3787 | 0.3617 | 0.2131 | 0.1166 | 0.1083 | 0.0017 | **0.3085** |
| Hispanic | 0.6148 | 0.4954 | 0.5367 | 0.2056 | 0.4266 | 0.3464 | 0.1827 | 0.0718 | 0.1074 | 0.1219 | **0.3109** |
| Woman | 0.4396 | 0.5065 | 0.7702 | 0.4978 | 0.2697 | 0.2842 | 0.1570 | 0.0821 | 0.1276 | 0.0090 | **0.3144** |
| 31~45 | 0.6628 | 0.5464 | 0.4344 | 0.4306 | 0.3909 | 0.3157 | 0.1866 | 0.1116 | 0.1312 | 0.0172 | **0.3227** |
| 46~60 | 0.4264 | 0.6078 | 0.5192 | 0.5523 | 0.4475 | 0.4217 | 0.1753 | 0.0600 | 0.0821 | 0.0020 | **0.3294** |
| Never discuss | 0.6939 | 0.7437 | 0.3298 | 0.0301 | 0.4017 | 0.5551 | 0.1653 | 0.3795 | 0.0239 | 0.0894 | **0.3413** |
| Asian | 0.8101 | 0.5723 | 0.3700 | 0.3542 | 0.3352 | 0.3469 | 0.3897 | 0.1685 | 0.1193 | 0.0228 | **0.3489** |
| Interest | 0.4871 | 0.4537 | 0.5215 | 0.9041 | 0.4556 | 0.2667 | 0.2250 | 0.0393 | 0.1397 | 0.0066 | **0.3499** |
| Liberals | 0.7249 | 0.7149 | 0.6979 | 0.5460 | 0.0639 | 0.1934 | 0.2778 | 0.0253 | 0.2936 | 0.1847 | **0.3722** |
| Moderates | 0.8077 | 0.7160 | 0.5855 | 0.2523 | 0.6202 | 0.3352 | 0.1493 | 0.2754 | 0.0608 | 0.0845 | **0.3887** |
| Democrats | 0.7520 | 0.7890 | 1.2404 | 0.6740 | 0.0848 | 0.5298 | 0.5671 | 0.0177 | 0.5225 | 0.2983 | **0.5476** |
| Independents | 1.1484 | 1.1928 | 0.5581 | 0.3433 | 0.5305 | 0.4277 | 0.2768 | 0.9092 | 0.0716 | 0.3442 | **0.5803** |
| Black | 1.0861 | 1.0458 | 1.1418 | 0.5164 | 0.1415 | 1.0995 | 0.6392 | 0.0848 | 0.3065 | 0.2556 | **0.6317** |

Table 2: KL-divergence values from stratified sampling experiments for multiple-choice questions. Darker the green color, the higher is the replicability by random silicon sampling; the darker the red color, the lower is the replicability. Bold column on far right represents the average KL-divergence values for each group.

### 5.1 Stratified random silicon sampling

We conducted random silicon sampling for 23 subgroups within the ANES respondents.

Experimental results revealed significant differences in the response distribution between the random silicon samples and ANES respondents for conservatives, moderates, democrats, independents, republicans, church goers, men, and women, with p-values exceeding 0.05. Figure 2 shows that the replicability for the party ID groups (democrats, independents, and republicans) is relatively low. This can be attributed to the extreme voting choices of party supporters in random silicon sample. The Joe Biden and Donald Trump voting rates among democratic and republican supporters in random silicon sample were 99.96 % and 99.22 %, respectively, which are more extreme than those in the actual ANES groups. Further, the model struggled the most in replicating the response distribution of the independent group. Prior political science research proposed that the independent group was difficult to predict because of the conflicting nature of their two-party choices, low likelihood of voting, limited political knowledge, and minimal interest in politics (Klar and Krupnikov 2016; Magleby et al., 2011; Argyle et al., 2023). Thus, we speculate that GPT 3.5 heavily relies on party-support information when generating presidential election choices. However, random silicon sampling demonstrated high replicability for the overall ANES group, which implies that interactions among the eight variables allowed the model to generate responses that strongly correlated with the group's opinion. In essence, the model exhibits biases not only towards specific single demographic variables but also towards the combinations of these variables (Cheng et al., 2023a).

### 5.2 Down-sized random silicon sampling

We compare the similarity of response distributions from smaller-sized random silicon samples with the entire ANES response distribution to identify the appropriate sample size for the method. Figure 3 shows that replicability decreased starting from a sample size of 3 % (=163). At 2 % (=108), the random silicon sample incorrectly favored Donald Trump (51.58 %) over Joe Biden (48.42 %). A possible interpretation could be similar to the emphasis in social science research on possessing a minimum of 200 samples for survey sampling (MacCallum et al., 1999; Guadagnoli and Velicer, 1988; Fan et al., 1999), which suggests that at least 200 samples may be necessary to accurately mirror the opinions of the target group (in our case, ANES 2020 respondents). The minimum sample size serves as a threshold for ensuring that responses generated by random silicon sampling adequately represent the diversity and range of opinions present in a larger population. In our study, response replication was feasible with a sample size of 4% (217).

However, the required sample size may vary depending on the number and complexity of demographic variables used in the prediction (Aliaga and Ruilin, 2006; Black et al., 2000; Crissman et al., 2017). Although our results indicate that replicability is maintained in samples larger than a certain size (~200), researchers using random silicon sampling must empirically explore the appropriate sample size starting with 200 samples.

### 5.3 Random silicon sampling on various topics

We experimented with a random silicon sampling of multiple-choice questions on various topics. Figure 4 illustrates that the random silicon sample can replicate opinions only for questions regarding "income inequality" (p > 0.05, KL-divergence = 0.00096).

Among the ten questions, extreme voting tendencies were found in six questions, which excludes those with the lowest KL-divergence value (i.e., "gun control" and "income inequality"), and 5-point scale questions (i.e., "current economy" and "climate change"). For example, as a response to the random silicon sample for the question "Does the increasing number of people of many different races and ethnic groups in the United States make this country a better place to live, a worse place to live, or does it make no difference?," 90 % chose "1 Better." For the question "Do you think it is better, worse, or makes no difference for the family as a whole if the man works outside the home and the woman takes care of the home and family?," 96 % chose "3. Makes no difference." Further, similar patterns were observed in the responses for the other four questions. We speculate that this tendency reflects the GPT-3.5's bias towards "harmlessness," considering the sensitivity of the questions topics (Carpenter, 2020; Lee et al., 2019; Nawyn, 2019; Artz et al., 2022). Models such as GPT-3.5 and GPT 4.0 developed by OpenAI were trained to prioritize not promoting discrimination against specific groups such as race and gender and to approach sensitive topics in a manner that safeguards human well-being (Ouyang et al., 2022; OpenAI, 2023; Zhong et al., 2023; Cheng et al., 2023a). However, there is a risk in that the model can over-generalize for nonharmful responses (Ouyang et al., 2022). Therefore, the model's responses were influenced more by a tendency to provide innocuous answers than by conditioned demographic information, especially regarding gender, race, homosexuality, and sensitive issues such as drug addiction, health insurance, and refugees.

We conducted stratified experiments on multiple questions to ascertain whether this trend was more pronounced in certain demographic subgroups. Response tendencies for the questions of each group tends to choose "harmless" options; however, this trend was more extreme in the specific group. Table 2 lists that the white, republican, and male groups had lower KL-divergences, whereas the democrats, independents, and black groups had higher KL-divergences. ANES responses of the white and black groups to the question regarding racial diversity were respectively similar to 54 % and 55 % for "1. The better", whereas it was 87 % for whites and 98 % for black groups in the random silicon sample, with a more extreme trend in the black group. In addition, ANES responses from the white group showed a 70.43 % approval rate for same-sex marriages, which was higher than the 57.08 % observed for the black group. In a random silicon sample, the approval rate among the white group was 77.46 %, whereas it was significantly higher in the black group at 95.67 % (see Tables A8 to A17). Therefore, we assume that the model has a more extreme "harmlessness" bias for certain demographic subgroups, and we look forward to exploring this in future research.

## 6 Conclusion

We expanded the boundaries of using LLMs to aggregate and synthesize opinions from various human subgroups. Existing methods relied on individual-level demographic information to condition LLMs, which is a requirement that was restrictive for practical applications (Simmons and Hare, 2023). Our results suggested that it is feasible to survey the opinions of a sub-group using only their group-level demographic information. Although LLMs cannot entirely replace human respondents in surveys, their cost-effectiveness (generating 5441 synthetic responses for only $0.7 in an hour) makes them invaluable for preliminary testing. Therefore, our "random silicon sampling," which simulates target group opinions based on their demographic distribution, can greatly benefit social scientists in designing theoretical models before actual surveys.

Furthermore, our research investigated the limitations of LLMs as proxies for survey respondents as described in existing studies. We observed a tendency in that the model towards extreme choices among party supporters in the issue of presidential candidate selection. We also observed a bias in the model towards harmless responses in surveys on sensitive topics. Furthermore, we discovered that the extent of this "harmlessness" bias varies among different subgroups within the population. Therefore, it is crucial to thoroughly consider and explore the inherent biases of the model related to both domains of opinions under investigation and the demographic variables defining the group to effectively employ random silicon sampling. In future, we plan to identify points where the biases of the models towards specific groups and topics intersect. By appropriately adjusting these biases, we can develop a systematic framework capable of accurately simulating real human opinions.

## 7 Limitations

**Scope of Experiments** In our study, we used demographic information from the ANES data to determine whether a language model could simulate the response distribution of ANES respondents. Although ANES is a primary source for gauging the American public opinion (ANES 2021; Argyle et al., 2023), a potential risk is that the model may have learned the outcomes associated with the ANES data results. Therefore, future studies must validate the replicability of random silicon sampling methodology using different survey results. Additionally, the ANES is an American public opinion survey dataset, with 70% of the respondents being Caucasian. The replicability of this methodology requires verification in diverse groups encompassing various national and ethnic backgrounds.

**API Calling Point** Our research involved conducting experiments using the OpenAI's GPT-3.5-turbo-0613. The changes in the training data and internal updates of the model can skew the experimental outcomes. Researchers aiming to generate survey responses using language models should be mindful of this issue.

**Inherent Biases in Language Models** Our experiments revealed that GPT-3.5 exhibits biases in responses related to specific groups and topics. The model showed an extreme tendency to provide "harmless" responses when answering questions about politically sensitive subjects. This tendency varied in intensity among different subgroups of the population, with the model displaying a stronger inclination towards extremely harmless responses for blacks compared to that for whites, and for democratic supporters over republican supporters.

We hope future research will further investigate and explore reasons why the model exhibits biases in predicting harmless opinions in certain subgroups.

## 8 Ethical Considerations

This study discussed the potential of using language models for generating survey responses. However, the possibilities that we presented may be misused.

1) If language models can generate human-like survey responses (especially at a low cost), there is a risk that online survey participants may use them to substitute for their own responses (Jansen et al., 2023; Hämäläinen et al., 2023).

2) Our methodology presents a more practical approach for generating synthetic opinions of subgroups using language models. However, this approach carries the risk of misuse, as one can generate synthetic responses that may be misconstrued as the actual opinions of a group. Further, as indicated by our results, LLMs still harbor biases that are not fully understood. Therefore, it is crucial to be aware that opinions generated using language models can be distorted because of these biases.

## A  Experimental Details

### A.1  ANES Demographic Data

We utilized the demographic information and responses of participants from the ANES 2020 pre-election dataset. We employed eight demographic variables, with the details of the questions used to gather each demographic information depicted in Table A1.

We used the response distributions for demographic questions to construct synthetic respondents, each with eight randomly selected demographic variables (e.g., V201549x:1, V202022:2, V201200: 6, ...). Each demographic variable is described in the first person and entered as a system prompt into the GPT 3.5 API. For the missing values in the ANES responses, such as non-responses or unwillingness to respond, we did not generate prompts. We used prompts of the same format as employed in the experiments by Argyle et al. (2023) to compare random silicon sampling and silicon sampling. Table A2 lists an example of these prompts.

We use the default hyperparameters (temperature = 1, top P = 1, frequency penalty = 0, presence penalty = 0, best of = 1). GPT-3.5 generations were produced from September 20th until November 14th, using the 2023-06-13-preview version of the API.

### A.2  Survey Data

In our initial experiment, we used voting choice questions about the 2020 U.S. presidential candidates. The questions are entered as user prompts into the GPT 3.5 API, following the demographic information of each synthetic respondent (Table A2). The generated response distributions and ANES response distributions were normalized similarly as Biden vs. Trump. In the ANES response distribution, nonresponses, refusals to respond, and votes for third and fourth candidates were treated as the missing values. In the generated response distribution, responses not referring to Biden or Trump were treated as missing values.

### A.3  Multiple Questions

We selected questions from the ANES survey on ten politically sensitive issues (Hwang et al., 2023; Kalinin, 2023; Feng et al., 2023). Table A3 describes the questions and answer choices for each item. The questions and choices for the

| Variable | Code | Answer choices |
|---|---|---|
| Race | V201549x | 1: 'white', 2: 'black', 3: 'asian', 4: 'native American', 5: 'hispanic' |
| Gender | V201600 | 1: "man", 2: "woman" |
| Age | V201507x | <Open-ended> |
| Ideology | V201200 | 1: "extremely liberal", 2: "liberal", 3: "slightly liberal", 4: "moderate", 5: "slightly conservative", 6: "conservative", 7: "extremely conservative" |
| Party ID | V201231x | 1: "a strong democrat", 2: "a weak Democrat", 3: "an independent who leans Democratic", 4: "an independent", 5: "an independent who leans Republican", 6: "a weak Republican", 7: "a strong Republican" |
| Political interest | V202406 | 1: "very", 2: "somewhat", 3: "not very", 4: "not at all" |
| Church attendance | V201452 | 1: "attend church", 2: "do not attend church" |
| Political discuss | V202022 | 1: 'I like to discuss politics with my family and friends.', 2: 'I never discuss politics with my family or friends.' |

Table A1: Demographic Variables of ANES 2020 Respondents. 'Code' refers to the variable names used in the ANES dataset.

---

**Respondent 1:**

**System prompt**: *Racially, I am black. I like to discuss politics with my family and friends. Ideologically, I am strongly liberal. Politically, I am a democrat. I do not attend church. I am 33 years old. I am a man. I am highly interested in politics.*

**User prompt**: *In the 2020 presidential election, Donald Trump is the Republican candidate, and Joe Biden is the Democratic candidate, and I voted for __ __*

**Respondent 2:**

**System prompt**: *Racially, I am white. I like to discuss politics with my family and friends. Ideologically, I am slightly conservative. Politically, I am a strong Republican. I attend church. I am 80 years old. I am a woman. I am somewhat interested in politics.*

**User prompt**: *In the 2020 presidential election, Donald Trump is the Republican candidate, and Joe Biden is the Democratic candidate, and I voted for __ __*

---

Table A2: Examples of Prompts. The underlined sections refer to randomly selected demographic variables for each synthetic respondent.

multiple questions were entered as user prompts. Some questions in the survey, such as those requiring a specific time reference like "currently" or "these days," necessitated the inclusion of a date description in the system prompt to mirror the survey date of ANES 2020. Table A4 illustrates an example of these prompts.

## B Stability Analysis

Our experiments found that we could generate highly similar response distributions with real ANES responses for the presidential questions using random silicon samples. Additional experiments were conducted to verify the stability of this replicability.

**Prompt Sensitivity** The generative capabilities of language models are highly sensitive to the input prompts, especially in the context of survey question responses, and they can be as sensitive to the order of questions as humans (Kalinin, 2023). Therefore, we conducted experiments by changing the order of Biden and Trump in the user prompts. Table A5 indicates that, even with altered prompts, the model can still generate highly similar response distributions with ANES respondents.

**Year Sensitivity** We conducted experiments with ANES data from the years 2012 and 2016. Using the ANES 2012 data, generated responses such as [*Barack Obama, Barack, Obama, the Democratic, a Democratic*] were coded as vote for "Barack

| Topic | Code | Question | Answer choices |
|---|---|---|---|
| Race diversity | V202371 | Does the increasing number of people of many different races and ethnic groups in the United States make this country a better place to live, a worse place to live, or does it make no difference? | 1. Better 2. Worse 3. Makes no difference |
| Gender role | V202287 | Do you think it is better, worse, or makes no difference for the family as a whole if the man works outside the home and the woman takes care of the home and family? | 1. Better 2. Worse 3. Makes no difference |
| Current Economy | V201324 | What do you think about the state of the economy these days in the United States? | 1. Very good 2. Good 3. Neither good nor bad 4. Bad 5. Very bad |
| Drug addiction | V202348 | Do you think the federal government should be doing more about the opioid drug addiction issue, should be doing less, or is it currently doing the right amount? | 1. Should be doing more 2. Should be doing less 3. Is doing the right amount |
| Climate change | V202332 | How much, if at all, do you think climate change is currently affecting severe weather events or temperature patterns in the United States? | 1. Not at all 2. A little 3. A moderate amount 4. A lot 5. A great deal |
| Gay marriage | V201416 | Which comes closest to your view? You can just tell me the number of your choice. | 1. Gay and lesbian couples should be allowed to legally marry. 2. Gay and lesbian couples should be allowed to form civil unions but not legally marry. 3. There should be no legal recognition of gay or lesbian couples' relationship. |
| Refugee allowing | V202234 | Do you favor, oppose, or neither favor nor oppose allowing refugees who are fleeing war, persecution, or natural disasters in other countries to come to live in the U.S.? | 1. Favor 2. Oppose 3. Neither favor nor oppose |
| Health insurance | V202378 | Do you favor an increase, decrease, or no change in government spending to help people pay for health insurance when people cannot pay for it all themselves? | 1. Increase 2. Decrease 3. No change |
| Gun regulation | V202337 | Do you think the federal government should make it more difficult for people to buy a gun than it is now, make it easier for people to buy a gun, or keep these rules about the same as they are now? | 1. More difficult 2. Easier 3. Keep these rules about the same |
| Income inequality | V202257 | Do you favor, oppose, or neither favor nor oppose the government trying to reduce the difference in incomes between the richest and poorest households? | 1. Favor 2. Oppose 3. Neither favor nor oppose |

Table A3: Multiple-choice questions selected from ANES 2020. "Code" refers to the variable names used in the ANES dataset.

---

**Respondent 1:**

**System prompt**: *Today is November 3, 2020. Racially, I am white. I like to discuss politics with my family and friends. Ideologically, I am slightly conservative. Politically, I am an independent who leans Republican. I do not attend church. I am 53 years old. I am a woman. I am somewhat interested in politics.*

**User prompt**: *Question: What do you think about the state of the economy these days in the United States?*

*Answer choices:*

*1. Very good 2. Good 3. Neither good nor bad 4. Bad 5. Very bad*

*My answer is __*

---

Table A4: Examples of prompts in the multiple-choice questions experiments. The underlined sections refer to randomly selected demographic variables for each synthetic respondent.

Obama," and [*Mitt Romney, Mitt Romney, the Republican, a Republican*] were coded as vote for "Mitt Romney." Further, [*Hillary Clinton, Hillary, Clinton, the Democratic, a Democratic*] were coded as a vote for "Hillary Clinton," and [*Donald Trump, Donald, Trump, the Republican, a Republican*] were coded as vote for "Donald Trump." Table A6 illustrates that simulating opinions was relatively difficult for the year 2012, while it was very high for 2016.

**Model Comparison** We compared the replicability of each model. In this experiment, we used the GPT-3.5-turbo API. We also conducted

| Sample | Biden rate | Trump rate | Chi-squared value | KL-divergence |
|---|---|---|---|---|
| ANES 2020 | 0.5888 | 0.4118 | | |
| Reverse sample 1 | 0.5738 | 0.4262 | 1.7616 | 0.00042 |
| Reverse sample 2 | 0.5691 | 0.4309 | 3.1348 | 0.00075 |
| Reverse sample 3 | 0.5691 | 0.4309 | 3.1348 | 0.00075 |
| Reverse sample 4 | 0.5684 | 0.4316 | 3.3735 | 0.00080 |
| Reverse sample 5 | 0.5655 | 0.4345 | 4.2862* | 0.00106 |

Table A5: Experimental results on altering the order of candidates in prompts for presidential election voting choice questions.

| Sample | Obama rate | Romney rate | Chi-squared value | KL-divergence |
|---|---|---|---|---|
| ANES 2012 | 0.5960 | 0.4040 | | |
| Silicon sample | 0.6257 | 0.3743 | 8.5855* | 0.00186 |
| Random silicon sample 1 | 0.6185 | 0.3815 | 4.9557* | 0.00106 |
| Random silicon sample 2 | 0.5960 | 0.4040 | 10.5530* | 0.00225 |

(a) ANES 2012

| Sample | Clinton rate | Trump rate | Chi-squared value | KL-divergence |
|---|---|---|---|---|
| ANES 2016 | 0.5228 | 0.4772 | | |
| Silicon sample | 0.5254 | 0.4746 | 0.0298 | 0.00001 |
| Random silicon sample 1 | 0.5192 | 0.4808 | 0.0677 | 0.00003 |
| Random silicon sample 2 | 0.5196 | 0.4804 | 0.0492 | 0.00002 |

(b) ANES 2016

Table A6: Experimental results using ANES data for Each Year: (a) 2012, (b) 2016.

experiments with the GPT-4 API. Table A7 illustrates that there is no differences in the results between the two models. In terms of cost, the GPT-3.5-turbo took ~$0.7 and 1 h, while GPT-4 took ~$20 and 2 h and 30 min.

| Sample | Biden rate | Trump rate | Chi-squared value | KL-divergence |
|---|---|---|---|---|
| ANES 2020 | 0.5888 | 0.4118 | | |
| GPT-3.5-turbo | 0.5800 | 0.4200 | 0.5688 | 0.00014 |
| GPT-4 | 0.5691 | 0.4309 | 3.2060 | 0.00075 |

Table A7: Comparison of experimental results using GPT-3.5-Turbo and GPT-4. The experimental results for GPT-3.5-Turbo are from the random silicon sample 1 in Table 1.

## C MQ Hierarchical Experiment

In our multiple-choice questions experiments, we assumed that the generated responses reflected an inherent harmless bias in the model. We conducted a hierarchical experiment to compare how this bias varied among different demographic subgroups. The harmless bias for sensitive questions varies across different demographic subgroups. Tables A8–A17 compare the response distributions of each demographic subgroup for each question.

| Variable | ANES | | | | | Silicon | | | | | Chi-squared value | KL-divergence |
|---|---|---|---|---|---|---|---|---|---|---|---|---|
| | 1 | 2 | 3 | 4 | 5 | 1 | 2 | 3 | 4 | 5 | | |
| Men | 6.80 % | 25.70 % | 26.40 % | 28.59 % | 12.51 % | 12.26 % | 18.04 % | 49.30 % | 20.10 % | 0.29 % | 550.6956* | 0.4554 |
| Women | 3.42 % | 20.98 % | 28.77 % | 33.04 % | 13.80 % | 6.98 % | 13.60 % | 53.28 % | 25.86 % | 0.28 % | 683.5539* | 0.5065 |
| Whites | 5.78 % | 26.46 % | 26.26 % | 30.03 % | 11.47 % | 12.11 % | 18.28 % | 48.99 % | 20.11 % | 0.50 % | 861.4956* | 0.3702 |
| Blacks | 1.65 % | 8.02 % | 30.66 % | 34.57 % | 25.10 % | 0.64 % | 2.75 % | 43.43 % | 52.97 % | 0.21 % | 159.0795* | 1.0458 |
| Asians | 4.38 % | 18.16 % | 34.66 % | 29.44 % | 13.36 % | 4.56 % | 10.20 % | 67.25 % | 17.79 % | 0.22 % | 131.8784* | 0.5723 |
| Native Americans | 2.03 % | 15.23 % | 30.96 % | 42.13 % | 9.64 % | 3.76 % | 6.99 % | 65.05 % | 23.12 % | 1.08 % | 53.5079* | 0.3407 |
| Hispanics | 3.48 % | 20.87 % | 29.57 % | 30.43 % | 15.65 % | 4.42 % | 11.50 % | 65.49 % | 17.70 % | 0.88 % | 37.4830* | 0.4954 |
| 18–30 years old | 3.70 % | 17.08 % | 32.26 % | 32.38 % | 14.58 % | 4.99 % | 11.94 % | 56.15 % | 26.43 % | 0.49 % | 176.0423* | 0.4324 |
| 31–45 years old | 4.48 % | 19.73 % | 28.62 % | 32.86 % | 14.31 % | 5.92 % | 16.49 % | 58.53 % | 18.73 % | 0.32 % | 351.7307* | 0.5464 |
| 46–60 years old | 4.78 % | 24.39 % | 25.96 % | 30.90 % | 13.96 % | 9.19 % | 18.55 % | 51.69 % | 20.40 % | 0.16 % | 329.4409* | 0.6078 |
| Over 60 | 5.86 % | 27.02 % | 25.52 % | 29.76 % | 11.84 % | 13.81 % | 15.63 % | 40.53 % | 29.03 % | 1.00 % | 324.6537* | 0.2792 |
| Liberals | 0.85 % | 8.33 % | 22.02 % | 45.44 % | 23.36 % | 0.57 % | 3.17 % | 31.09 % | 64.59 % | 0.57 % | 458.9957* | 0.7149 |
| Moderates | 2.86 % | 17.16 % | 35.13 % | 32.92 % | 11.93 % | 2.36 % | 9.54 % | 77.55 % | 10.38 % | 0.17 % | 495.8219* | 0.7160 |
| Conservatives | 11.37 % | 44.95 % | 23.72 % | 16.73 % | 3.23 % | 31.82 % | 33.00 % | 27.38 % | 7.51 % | 0.30 % | 303.7140* | 0.1992 |
| Democrats | 1.11 % | 8.65 % | 25.40 % | 42.67 % | 22.17 % | 0.53 % | 3.57 % | 39.10 % | 56.47 % | 0.33 % | 693.5930* | 0.7890 |
| Independent | 1.69 % | 14.62 % | 38.92 % | 31.69 % | 13.08 % | 0.16 % | 1.42 % | 93.22 % | 5.21 % | 0.00 % | 424.9014* | 1.1928 |
| Republicans | 10.41 % | 42.29 % | 26.92 % | 17.40 % | 2.98 % | 27.69 % | 34.34 % | 32.81 % | 4.89 % | 0.28 % | 403.0588* | 0.2244 |
| Attends church | 6.17 % | 29.56 % | 27.37 % | 26.86 % | 10.04 % | 15.67 % | 20.66 % | 51.55 % | 11.72 % | 0.40 % | 702.1086* | 0.4206 |
| Does not attend church | 3.90 % | 17.42 % | 28.08 % | 34.56 % | 16.04 % | 3.90 % | 11.92 % | 48.33 % | 35.34 % | 0.51 % | 574.5026* | 0.4576 |
| Discusses politics | 4.81 % | 23.35 % | 26.69 % | 31.93 % | 13.23 % | 8.98 % | 15.41 % | 51.03 % | 24.22 % | 0.36 % | 1012.1402* | 0.4603 |
| Does not discuss politics | 5.05 % | 20.20 % | 38.64 % | 23.99 % | 12.12 % | 1.02 % | 8.44 % | 84.14 % | 6.39 % | 0.00 % | 183.2899* | 0.7437 |
| Somewhat interested in politics | 4.98 % | 23.78 % | 24.88 % | 32.17 % | 14.19 % | 10.48 % | 17.16 % | 44.17 % | 27.79 % | 0.39 % | 727.9616* | 0.4537 |
| Not very interested in politics | 4.29 % | 21.37 % | 34.63 % | 29.41 % | 10.30 % | 4.95 % | 13.01 % | 75.42 % | 6.23 % | 0.40 % | 523.2719* | 0.6217 |

Table A8: Results of the stratified experiment on multiple-choice questions about **current economy**. Columns "1, 2, 3, 4, and "5" represent the proportions of answer choices in the sample. Chi-squared test p-value <0.05 with * indicates that there is statistically significant difference between the response distributions of the sample and ANES.

| Variable | ANES | | | Silicon | | | Chi-squared value | KL-divergence |
|---|---|---|---|---|---|---|---|---|
| | 1 | 2 | 3 | 1 | 2 | 3 | | |
| Men | 67.33 % | 19.05 % | 13.62 % | 76.44 % | 22.56 % | 1.00 % | 283.7214* | 0.2384 |
| Women | 71.44 % | 16.50 % | 12.06 % | 85.01 % | 14.51 % | 0.49 % | 345.3145* | 0.2842 |
| Whites | 70.32 % | 17.75 % | 11.93 % | 77.46 % | 21.68 % | 0.86 % | 395.4352* | 0.2101 |
| Blacks | 57.08 % | 19.79 % | 23.13 % | 95.67 % | 4.33 % | 0.00 % | 207.1025* | 1.0995 |
| Asians | 73.84 % | 14.98 % | 11.18 % | 87.13 % | 12.66 % | 0.21 % | 56.1996* | 0.3469 |
| Native Americans | 74.23 % | 18.04 % | 7.73 % | 89.85 % | 10.15 % | 0.00 % | 22.4617* | 0.1776 |
| Hispanics | 65.22 % | 18.26 % | 16.52 % | 81.58 % | 17.54 % | 0.88 % | 18.1489* | 0.3464 |
| 18–30 years old | 80.96 % | 9.94 % | 9.10 % | 91.28 % | 8.72 % | 0.00 % | 80.9888* | 0.3103 |
| 31–45 years old | 76.28 % | 13.87 % | 9.85 % | 86.33 % | 13.51 % | 0.16 % | 125.9635* | 0.3157 |
| 46–60 years old | 67.64 % | 18.86 % | 13.50 % | 79.95 % | 19.81 % | 0.24 % | 173.4504* | 0.4217 |
| Over 60 | 61.69 % | 22.88 % | 15.42 % | 73.05 % | 24.54 % | 2.41 % | 183.2982* | 0.1659 |
| Liberals | 92.39 % | 4.69 % | 2.92 % | 99.81 % | 0.19 % | 0.00 % | 119.3462* | 0.1934 |
| Moderates | 76.41% | 14.90 % | 8.69 % | 89.11 % | 10.81 % | 0.08 % | 122.0578* | 0.3352 |
| Conservatives | 45.87 % | 32.19 % | 21.94 % | 37.04 % | 58.88 % | 4.08 % | 357.2579* | 0.2727 |
| Democrats | 85.53 % | 8.45 % | 6.03 % | 99.88 % | 0.12 % | 0.00 % | 382.2564* | 0.5298 |
| Independent | 70.98 % | 16.40 % | 12.62 % | 93.30 % | 6.39 % | 0.31 % | 122.6862* | 0.4277 |
| Republicans | 50.86 % | 28.48 % | 20.66 % | 39.55 % | 56.78 % | 3.67 % | 492.1167* | 0.2884 |
| Attend church | 54.96 % | 25.21 % | 19.83 % | 68.74 % | 29.87 % | 1.39 % | 453.0548* | 0.3617 |
| Does not attend church | 82.78 % | 10.82 % | 6.41 % | 89.14 % | 10.68 % | 0.18 % | 170.2753* | 0.1689 |
| Discusses politics | 70.21 % | 17.62 % | 12.17 % | 81.09 % | 18.21 % | 0.70 % | 472.1950* | 0.2405 |
| Does not discuss politics | 64.87 % | 18.46 % | 16.67 % | 72.86 % | 27.14 % | 0.00 % | 74.6477* | 0.5551 |
| Somewhat interested in politics | 69.98 % | 18.31 % | 11.71 % | 82.40 % | 17.10 % | 0.50 % | 383.3780* | 0.2667 |
| Not very interested in politics | 69.18 % | 16.01 % | 14.82 % | 77.98 % | 21.55 % | 0.47 % | 187.4560* | 0.3798 |

Table A9: Results of the stratified experiment on multiple-choice questions about **gay marriage**. Columns "1, 2, and 3" represent the proportions of answer choices in the sample. Chi-squared test p-value <0.05 with * indicates that there is a statistically significant difference between the response distributions of the sample and ANES.

| Variable | ANES | | | Silicon | | | Chi-squared value | KL-divergence |
|---|---|---|---|---|---|---|---|---|
| | 1 | 2 | 3 | 1 | 2 | 3 | | |
| Men | 52.57 % | 15.36 % | 32.07 % | 72.48 % | 3.43 % | 24.09 % | 269.3664* | 0.1532 |
| Women | 56.09 % | 12.58 % | 31.33 % | 77.43 % | 2.65 % | 19.92 % | 338.0146* | 0.1570 |
| Whites | 54.95 % | 13.73 % | 31.32 % | 73.31 % | 4.30 % | 22.39 % | 324.3213* | 0.1061 |
| Blacks | 49.05 % | 16.59 % | 34.36 % | 88.10 % | 0.63 % | 11.27 % | 173.6850* | 0.6392 |
| Asians | 56.47 % | 11.44 % | 32.09 % | 76.92 % | 0.21 % | 22.86 % | 70.6704* | 0.3897 |
| Native Americans | 56.21 % | 10.65 % | 33.14 % | 77.84 % | 1.08 % | 21.08 % | 25.2165* | 0.2106 |
| Hispanics | 45.45 % | 16.16 % | 38.38 % | 63.30 % | 2.75 % | 33.94 % | 13.5112* | 0.1827 |
| 18–30 years old | 65.07 % | 10.14 % | 24.79 % | 74.75 % | 1.59 % | 23.65 % | 55.0717* | 0.1090 |
| 31–45 years old | 57.31 % | 12.82 % | 29.87 % | 75.57 % | 1.63 % | 22.80 % | 145.2783* | 0.1866 |
| 46–60 years old | 50.18 % | 16.79 % | 33.03 % | 70.97 % | 3.47 % | 25.56 % | 156.4223* | 0.1753 |
| Over 60 | 51.68 % | 13.82 % | 34.50 % | 78.06 % | 5.25 % | 16.70 % | 253.7905* | 0.1711 |
| Liberals | 82.88 % | 3.87 % | 13.25 % | 98.09 % | 0.00 % | 1.91 % | 218.2147* | 0.2778 |
| Moderates | 51.50 % | 11.26 % | 37.24 % | 58.69 % | 1.30 % | 40.02 % | 96.9313* | 0.1493 |
| Conservatives | 35.77 % | 23.76 % | 40.48 % | 61.67 % | 12.42 % | 25.91 % | 211.2294* | 0.1398 |
| Democrats | 73.62 % | 5.89 % | 20.49 % | 98.07 % | 0.00 % | 1.93 % | 605.9621* | 0.5671 |
| Independent | 44.48 % | 11.73 % | 43.78 % | 35.12 % | 0.65 % | 64.23 % | 90.0622* | 0.2768 |
| Republicans | 35.34 % | 23.65 % | 41.01 % | 59.41 % | 11.49 % | 29.10 % | 245.8289* | 0.1278 |
| Attend church | 49.80 % | 15.46 % | 34.74 % | 73.40 % | 2.55 % | 24.04 % | 372.5310* | 0.2131 |
| Does not attend church | 58.64 % | 12.33 % | 29.03 % | 78.31 % | 2.30 % | 19.39 % | 304.4038* | 0.1544 |
| Discusses politics | 56.10 % | 13.41 % | 30.49 % | 77.34 % | 2.84 % | 19.82 % | 528.8234* | 0.1596 |
| Does not discuss politics | 36.11 % | 17.93 % | 45.96 % | 46.68 % | 3.45 % | 49.87 % | 43.1177* | 0.1653 |
| Somewhat interested in politics | 57.46 % | 14.16 % | 28.38 % | 82.71 % | 2.40 % | 14.88 % | 569.7230* | 0.2250 |
| Not very interested in politics | 46.37 % | 12.96 % | 40.67 % | 54.15 % | 4.73 % | 41.11 % | 54.3094* | 0.0541 |

Table A10: Results of the stratified experiment on multiple-choice questions about **allowing refugees**. Columns "1, 2, and 3" represent the proportions of answer choices in the sample. A chi-squared test p-value <0.05 with * indicates that there is a statistically significant difference between the response distributions of the sample and ANES.

| Variable | ANES | | | Silicon | | | Chi-squared value | KL-divergence |
|---|---|---|---|---|---|---|---|---|
| | 1 | 2 | 3 | 1 | 2 | 3 | | |
| Men | 46.04 % | 31.04 % | 22.93 % | 44.09 % | 29.15 % | 26.76 % | 8.9927* | 0.0039 |
| Women | 46.97 % | 26.05 % | 26.98 % | 52.54 % | 20.87 % | 26.58 % | 24.2053* | 0.0090 |
| Whites | 44.90 % | 31.11 % | 23.99 % | 44.98 % | 31.15 % | 23.87 % | 0.0148 | 0.0000 |
| Blacks | 53.68 % | 20.19 % | 26.13 % | 76.91 % | 3.09 % | 20.00 % | 81.7787* | 0.2556 |
| Asians | 48.25 % | 21.75 % | 30.00 % | 53.09 % | 13.86 % | 33.05 % | 9.3138 * | 0.0228 |
| Native Americans | 56.29 % | 13.77 % | 29.94 % | 60.82 % | 9.28 % | 29.90 % | 1.9106 | 0.0112 |
| Hispanics | 41.84 % | 28.57 % | 29.59 % | 34.21 % | 14.04 % | 51.75 % | 12.4132* | 0.1219 |
| 18–30 years old | 54.25 % | 19.83 % | 25.92 % | 50.72 % | 16.11 % | 33.17 % | 10.6127* | 0.0138 |
| 31–45 years old | 49.28 % | 25.32 % | 25.41 % | 49.76 % | 18.65 % | 31.59 % | 19.9266* | 0.0172 |
| 46–60 years old | 44.75 % | 29.24 % | 26.01 % | 47.34 % | 26.61 % | 26.05 % | 2.3257 | 0.0020 |
| Over 60 | 43.63 % | 33.00 % | 23.38 % | 50.31 % | 31.35 % | 18.34 % | 18.9574* | 0.0114 |
| Liberals | 77.86 % | 7.29 % | 14.85 % | 94.48 % | 0.74 % | 4.79 % | 192.8562* | 0.1847 |
| Moderates | 47.56 % | 19.36 % | 33.08 % | 39.83 % | 9.26 % | 50.91 % | 91.3720 * | 0.0845 |
| Conservatives | 17.83 % | 58.05 % | 24.12 % | 11.40 % | 77.09 % | 11.52 % | 138.3295* | 0.0935 |
| Democrats | 70.41 % | 10.06 % | 19.53 % | 92.06 % | 0.52 % | 7.43 % | 412.9242 * | 0.2983 |
| Independent | 42.73 % | 16.29 % | 40.98 % | 16.38 % | 4.68 % | 78.94 % | 184.1792 * | 0.3442 |
| Republicans | 20.16 % | 53.09 % | 26.75 % | 10.33 % | 74.66 % | 15.01 % | 208.2806 * | 0.1083 |
| Attend church | 38.82 % | 35.83 % | 25.35 % | 41.59 % | 34.73 % | 23.67 % | 4.0285 | 0.0017 |
| Does not attend church | 53.57 % | 21.63 % | 24.80 % | 58.51 % | 16.88 % | 24.61 % | 21.1768* | 0.0083 |
| Discusses politics | 47.94 % | 28.76 % | 23.30 % | 50.31 % | 25.61 % | 24.08 % | 10.9458 * | 0.0026 |
| Does not discuss politics | 30.89 % | 24.30 % | 44.81 % | 20.45 % | 14.14 % | 65.40 % | 34.2279 * | 0.0894 |
| Somewhat interested in politics | 49.59 % | 29.13 % | 21.28 % | 55.31 % | 25.77 % | 18.92 % | 22.3888 * | 0.0066 |
| Not very interested in politics | 38.34 % | 26.45 % | 35.21 % | 26.78 % | 23.77 % | 49.45 % | 58.2582 * | 0.0462 |

Table A11: Results of the stratified experiment on multiple-choice questions about **income inequality**. Columns "1, 2, and 3" represent the proportions of answer choices in the sample. Chi-squared test p-value <0.05 with * indicates that there is statistically significant difference between the response distributions of the sample and ANES.

| Variable | ANES | | | Silicon | | | Chi-squared value | KL-divergence |
|---|---|---|---|---|---|---|---|---|
| | 1 | 2 | 3 | 1 | 2 | 3 | | |
| Men | 32.13 % | 5.27 % | 62.61 % | 4.97 % | 0.28 % | 94.76 % | 669.4725* | 0.4956 |
| Women | 27.72 % | 5.47 % | 66.81 % | 4.11 % | 0.36 % | 95.53 % | 686.9062 * | 0.4396 |
| Whites | 30.12 % | 4.20 % | 65.67 % | 5.87 % | 0.47 % | 93.67 % | 831.2568 * | 0.3521 |
| Blacks | 30.95 % | 8.10 % | 60.95 % | 0.68 % | 0.68 % | 98.63 % | 191.9738 * | 1.0861 |
| Asians | 30.00 % | 7.75 % | 62.25 % | 1.93 % | 0.24 % | 97.83 % | 163.1435 * | 0.8101 |
| Native Americans | 22.49 % | 10.65 % | 66.86 % | 1.17 % | 0.00 % | 98.83 % | 61.5109 * | 0.7184 |
| Hispanics | 33.33 % | 8.08 % | 58.59 % | 3.81 % | 0.00 % | 96.19 % | 42.2187 * | 0.6148 |
| 18–30 years old | 19.69 % | 9.92 % | 70.40 % | 1.06 % | 0.13 % | 98.81 % | 231.9906* | 0.7642 |
| 31–45 years old | 24.48 % | 7.59 % | 67.93 % | 1.91 % | 0.17 % | 97.92 % | 364.2417 * | 0.6628 |
| 46–60 years old | 30.27 % | 4.14 % | 65.59 % | 4.65 % | 0.35 % | 95.00 % | 310.4081 * | 0.4264 |
| Over 60 | 37.32 % | 2.77 % | 59.91 % | 8.87 % | 0.93 % | 90.20 % | 374.4857 * | 0.3213 |
| Liberals | 13.42 % | 8.19 % | 78.39 % | 0.28 % | 0.07 % | 99.66 % | 336.2629 * | 0.7249 |
| Moderates | 28.67 % | 4.42 % | 66.92 % | 0.99 % | 0.45 % | 98.57 % | 389.1739 * | 0.8077 |
| Conservatives | 44.47 % | 3.00 % | 52.53 % | 21.93 % | 0.53 % | 77.54 % | 212.3627 * | 0.1616 |
| Democrats | 19.28 % | 7.12 % | 73.59 % | 0.45 % | 0.22 % | 99.33 % | 632.5292 * | 0.7520 |
| Independent | 27.23 % | 6.11 % | 66.67 % | 0.34 % | 0.17 % | 99.49 % | 226.8010 * | 1.1484 |
| Republicans | 42.43 % | 3.08 % | 54.49 % | 17.63 % | 0.97 % | 81.39 % | 322.7734 * | 0.1894 |
| Attend church | 36.60 % | 4.21 % | 59.19 % | 9.94 % | 1.33 % | 88.73 % | 498.4360 * | 0.2859 |
| Does not attend church | 23.53 % | 6.40 % | 70.07 % | 1.39 % | 0.35 % | 98.26 % | 765.7254 * | 0.6150 |
| Discusses politics | 29.88 % | 5.23 % | 64.88 % | 3.70 % | 0.26 % | 96.04 % | 1216.3296* | 0.5271 |
| Does not discuss politics | 27.85 % | 6.58 % | 65.57 % | 1.62 % | 0.54 % | 97.84 % | 130.2186 * | 0.6939 |
| Somewhat interested in politics | 29.80 % | 5.35 % | 64.84 % | 4.31 % | 0.26 % | 95.43 % | 918.3677 * | 0.4871 |
| Not very interested in politics | 29.51 % | 5.31 % | 65.18 % | 5.21 % | 0.78 % | 94.01 % | 301.8231 * | 0.3745 |

Table A12: Results of the stratified experiment on the multiple-choice questions about **gender roles**. Columns "1, 2', and 3" represent the proportions of answer choices in the sample. Chi-squared test p-values <0.05 with * indicate that there is statistically significant difference between the response distributions of the sample and ANES.

| Variable | ANES | | | | | Silicon | | | | | Chi-squared value | KL-divergence |
|---|---|---|---|---|---|---|---|---|---|---|---|---|
| | 1 | 2 | 3 | 4 | 5 | 1 | 2 | 3 | 4 | 5 | | |
| Men | 10.65 % | 17.46 % | 19.33 % | 18.16 % | 34.40 % | 0.72 % | 15.86 % | 53.00 % | 7.93 % | 22.49 % | 673.3515* | 0.4052 |
| Women | 7.84 % | 12.98 % | 21.64 % | 17.58 % | 39.96 % | 0.53 % | 13.39 % | 47.54 % | 10.07 % | 28.46 % | 520.5569 * | 0.2697 |
| Whites | 10.09 % | 16.93 % | 20.07 % | 16.75 % | 36.15 % | 0.96 % | 17.13 % | 53.10 % | 7.53 % | 21.29 % | 1024.9799* | 0.3658 |
| Blacks | 5.76 % | 6.95 % | 24.70 % | 20.38 % | 42.21 % | 0.00 % | 6.36 % | 30.91 % | 17.27 % | 45.45 % | 30.0134 * | 0.1415 |
| Asians | 6.50 % | 10.75 % | 20.50 % | 22.25 % | 40.00 % | 0.67 % | 9.62 % | 55.26 % | 9.40 % | 25.06 % | 124.0994* | 0.3352 |
| Native Americans | 2.98 % | 5.36 % | 22.62 % | 25.00 % | 44.05 % | 0.00 % | 9.39 % | 49.72 % | 11.05 % | 29.83 % | 39.0880 * | 0.2331 |
| Hispanics | 14.43 % | 13.40 % | 23.71 % | 14.43 % | 34.02 % | 0.93 % | 14.95 % | 53.27 % | 5.61 % | 25.23 % | 29.4075 * | 0.4266 |
| 18–30 years old | 6.08 % | 11.88 % | 22.21 % | 19.66 % | 40.17 % | 0.39 % | 14.73 % | 49.35 % | 9.30 % | 26.23 % | 165.3064* | 0.2829 |
| 31–45 years old | 7.99 % | 14.97 % | 23.23 % | 18.97 % | 34.85 % | 0.25 % | 15.21 % | 51.93 % | 6.47 % | 26.13 % | 295.7193 * | 0.3909 |
| 46–60 years old | 10.45 % | 16.04 % | 20.27 % | 16.85 % | 36.40 % | 0.35 % | 14.52 % | 52.98 % | 9.08 % | 23.08 % | 334.6376 * | 0.4475 |
| Over 60 | 10.23 % | 14.72 % | 18.28 % | 17.16 % | 39.61 % | 0.80 % | 12.54 % | 50.57 % | 9.07 % | 27.02 % | 437.3192 * | 0.3591 |
| Liberals | 0.89 % | 2.52 % | 8.87 % | 20.26 % | 67.46 % | 0.00 % | 1.04 % | 19.32 % | 15.91 % | 63.72 % | 87.2613 * | 0.0639 |
| Moderates | 3.85 % | 10.80 % | 22.63 % | 24.13 % | 38.59 % | 0.09 % | 16.48 % | 64.68 % | 7.62 % | 11.13 % | 539.9425* | 0.6202 |
| Conservatives | 22.01 % | 30.75 % | 27.15 % | 10.94 % | 9.14 % | 2.46 % | 33.94 % | 55.59 % | 3.88 % | 4.14 % | 471.9732 * | 0.4438 |
| Democrats | 1.31 % | 3.25 % | 11.75 % | 21.37 % | 62.31 % | 0.09 % | 1.70 % | 25.31 % | 15.95 % | 56.94 % | 167.6188 * | 0.0848 |
| Independent | 4.90 % | 14.19 % | 31.00 % | 19.09 % | 30.82 % | 0.50 % | 23.00 % | 64.17 % | 3.33 % | 9.00 % | 237.5227 * | 0.5305 |
| Republicans | 19.52 % | 28.79 % | 27.75 % | 13.33 % | 10.62 % | 3.16 % | 32.21 % | 58.96 % | 2.91 % | 2.76 % | 679.6660 * | 0.4598 |
| Attend church | 11.58 % | 18.74 % | 22.54 % | 16.37 % | 30.77 % | 1.18 % | 20.65 % | 54.84 % | 7.23 % | 16.09 % | 682.5612 * | 0.3787 |
| Does not attend church | 7.03 % | 11.66 % | 18.81 % | 19.06 % | 43.44 % | 0.31 % | 10.85 % | 45.10 % | 11.25 % | 32.48 % | 515.5345 * | 0.2890 |
| Discusses politics | 9.10 % | 15.19 % | 19.21 % | 17.87 % | 38.64 % | 0.77 % | 14.38 % | 51.52 % | 9.47 % | 23.85 % | 1141.6051* | 0.3429 |
| Does not discuss politics | 10.15 % | 12.94 % | 35.79 % | 17.26 % | 23.86 % | 1.87 % | 41.18 % | 42.78 % | 4.81 % | 9.36 % | 131.7435 * | 0.4017 |
| Somewhat interested in politics | 9.96 % | 14.41 % | 17.35 % | 16.77 % | 41.51 % | 0.30 % | 7.99 % | 50.69 % | 10.33 % | 30.69 % | 979.5666 * | 0.4556 |
| Not very interested in politics | 7.17 % | 16.59 % | 29.21 % | 20.56 % | 26.48 % | 1.18 % | 47.39 % | 41.93 % | 3.70 % | 5.80 % | 567.5017 * | 0.6046 |

Table A13: Results of the stratified experiment on multiple-choice questions about **climate change**. Columns "1, 2, 3, 4, and 5" represent the proportions of answer choices in the sample. Chi-squared test p-values <0.05 with * indicate that there is statistically significant difference between the response distributions of the sample and ANES.

| Variable | ANES | | | Silicon | | | Chi-squared value | KL-divergence |
|---|---|---|---|---|---|---|---|---|
| | 1 | 2 | 3 | 1 | 2 | 3 | | |
| Men | 46.24 % | 7.89 % | 45.87 % | 27.99 % | 7.78 % | 64.23 % | 168.1116* | 0.0788 |
| Women | 57.96 % | 4.88 % | 37.16 % | 39.74 % | 2.97 % | 57.29 % | 215.4035 * | 0.0821 |
| Whites | 50.07 % | 6.34 % | 43.58 % | 30.22 % | 7.10 % | 62.68 % | 297.4775 * | 0.0874 |
| Blacks | 65.47 % | 6.71 % | 27.82 % | 58.05 % | 1.27 % | 40.68 % | 29.7014 * | 0.0848 |
| Asians | 59.19 % | 6.05 % | 34.76 % | 35.82 % | 1.76 % | 62.42 % | 67.9033 * | 0.1685 |
| Native Americans | 68.67 % | 2.41 % | 28.92 % | 37.04 % | 1.59 % | 61.38 % | 37.5271 * | 0.2165 |
| Hispanics | 40.40 % | 11.11 % | 48.48 % | 30.28 % | 4.59 % | 65.14 % | 6.9018 * | 0.0718 |
| 18–30 years old | 56.17 % | 7.94 % | 35.89 % | 36.24 % | 2.70 % | 61.06 % | 101.6664* | 0.1411 |
| 31–45 years old | 52.63 % | 6.97 % | 40.40 % | 31.93 % | 4.37 % | 63.70 % | 125.8027* | 0.1116 |
| 46–60 years old | 49.28 % | 5.95 % | 44.77 % | 32.98 % | 5.59 % | 61.43 % | 68.0670 * | 0.0600 |
| Over 60 | 54.07 % | 5.26 % | 40.68 % | 39.01 % | 6.17 % | 54.81 % | 74.0526 * | 0.0467 |
| Liberals | 81.87 % | 2.18 % | 15.95 % | 85.47 % | 0.32% | 14.21 % | 24.5491 * | 0.0253 |
| Moderates | 55.55 % | 4.51 % | 39.94 % | 26.54 % | 0.43% | 73.03 % | 261.5824* | 0.2754 |
| Conservatives | 23.68 % | 11.27 % | 65.04 % | 8.03 % | 18.61 % | 73.36 % | 154.8963 * | 0.1214 |
| Democrats | 78.27 % | 2.21 % | 19.51 % | 79.36 % | 0.45 % | 20.19 % | 28.2662 * | 0.0177 |
| Independent | 46.75 % | 8.08 % | 45.17 % | 6.54 % | 0.16 % | 93.30 % | 333.7315* | 0.9092 |
| Republicans | 24.49 % | 10.58 % | 64.93 % | 6.71 % | 16.41 % | 76.88 % | 245.9727 * | 0.1609 |
| Attend church | 47.72 % | 6.99 % | 45.30 % | 25.84 % | 6.33 % | 67.82 % | 253.9644 * | 0.1166 |
| Does not attend church | 57.02 % | 5.77 % | 37.21 % | 44.38 % | 3.13 % | 52.49 % | 127.9568 * | 0.0502 |
| Discusses politics | 53.46 % | 5.96 % | 40.58 % | 37.54 % | 5.23 % | 57.24 % | 238.2273 * | 0.0573 |
| Does not discuss politics | 42.46 % | 10.23 % | 47.31 % | 16.20 % | 1.03 % | 82.78 % | 112.7974 * | 0.3795 |
| Somewhat interested in politics | 54.99 % | 5.82 % | 39.18 % | 42.12 % | 4.73 % | 53.15% | 131.1405 * | 0.0393 |
| Not very interested in politics | 46.09 % | 7.58 % | 46.33 % | 15.47 % | 6.43 % | 78.09 % | 292.3285 * | 0.2737 |

Table A14: Results of the stratified experiment on multiple-choice questions about **gun regulation**. Columns "1, 2, and 3" represent the proportions of answer choices in the sample. The chi-squared test p-values <0.05 with * indicate that there is a statistically significant difference between the response distributions of the sample and ANES.

| Variable | ANES | | | Silicon | | | Chi-squared value | KL-divergence |
|---|---|---|---|---|---|---|---|---|
| | 1 | 2 | 3 | 1 | 2 | 3 | | |
| Men | 69.38 % | 4.80 % | 25.81 % | 95.83 % | 0.63 % | 3.54 % | 569.2158* | 0.3864 |
| Women | 69.60 % | 4.32 % | 26.08 % | 97.25 % | 0.42 % | 2.33 % | 778.1386 * | 0.4978 |
| Whites | 68.99 % | 4.42 % | 26.59 % | 96.34 % | 0.65 % | 3.00 % | 978.9656 * | 0.4341 |
| Blacks | 71.50 % | 6.04 % | 22.46 % | 97.93 % | 0.00 % | 2.07 % | 128.0736 * | 0.5164 |
| Asians | 72.54 % | 4.79 % | 22.67 % | 96.38 % | 0.43 % | 3.19 % | 98.6269 * | 0.3542 |
| Native Americans | 71.26 % | 1.80 % | 26.95 % | 96.92 % | 1.54 % | 1.54 % | 50.7973 * | 0.5551 |
| Hispanics | 68.04 % | 6.19 % | 25.77 % | 91.15 % | 0.00 % | 8.85 % | 19.4229 * | 0.2056 |
| 18–30 years old | 73.97 % | 6.79 % | 19.24 % | 94.28 % | 0.73 % | 4.99 % | 124.6355* | 0.2316 |
| 31–45 years old | 68.82 % | 5.38 % | 25.80 % | 95.93 % | 0.24 % | 3.83 % | 310.2419 * | 0.4306 |
| 46–60 years old | 65.61 % | 4.17 % | 30.22 % | 96.65 % | 0.64 % | 2.72 % | 385.0402 * | 0.5523 |
| Over 60 | 70.19 % | 3.14 % | 26.67 % | 97.13 % | 0.86 % | 2.01% | 460.1168 * | 0.5019 |
| Liberals | 79.38 % | 2.33 % | 18.29 % | 99.45 % | 0.00 % | 0.55 % | 342.7998 * | 0.5460 |
| Moderates | 67.74 % | 3.78 % | 28.48 % | 92.24 % | 0.67 % | 7.10 % | 217.3349 * | 0.2523 |
| Conservatives | 61.46 % | 6.61 % | 31.93 % | 96.41 % | 1.18 % | 2.41 % | 611.2267 * | 0.6618 |
| Democrats | 78.94 % | 2.54 % | 18.52 % | 99.68 % | 0.00 % | 0.32 % | 559.8330 * | 0.6740 |
| Independent | 60.11 % | 6.62 % | 33.27 % | 89.01 % | 0.32 % | 10.67 % | 139.4058 * | 0.3433 |
| Republicans | 61.15 % | 6.32 % | 32.53 % | 94.17 % | 1.16 % | 4.67 % | 659.8825 * | 0.4745 |
| Attend church | 67.93 % | 4.54 % | 27.53 % | 97.57 % | 0.16 % | 2.27 % | 758.7729 * | 0.5934 |
| Does not attend church | 70.80 % | 4.58 % | 24.62 % | 96.94 % | 0.40 % | 2.66 % | 687.2063 * | 0.4376 |
| Discusses politics | 69.91 % | 4.51 % | 25.58 % | 97.41 % | 0.33 % | 2.26 % | 1187.3058* | 0.5070 |
| Does not discuss politics | 64.29 % | 5.36 % | 30.36 % | 72.55 % | 1.90 % | 25.54 % | 9.6195 * | 0.0301 |
| Somewhat interested in politics | 71.35 % | 4.01 % | 24.65 % | 99.44 % | 0.15 % | 0.41 % | 1076.6880 * | 0.9041 |
| Not very interested in politics | 64.59 % | 6.06 % | 29.35 % | 80.78 % | 1.54 % | 17.68 % | 91.9923 * | 0.0872 |

Table A15: Results of the stratified experiment on multiple-choice questions about **drug addiction**. Columns "1, 2, and 3" represent the proportions of answer choices in the sample. The chi-squared test p-values <0.05 with * indicate that there is a statistically significant difference between the response distributions of the sample and ANES.

| Variable | ANES | | | Silicon | | | Chi-squared value | KL-divergence |
|---|---|---|---|---|---|---|---|---|
| | 1 | 2 | 3 | 1 | 2 | 3 | | |
| Men | 55.10 % | 7.29 % | 37.61 % | 83.40 % | 0.34 % | 16.25 % | 467.0109* | 0.3100 |
| Women | 56.20 % | 6.30 % | 37.50 % | 94.71 % | 0.04 % | 5.25 % | 1112.2152* | 0.7702 |
| Whites | 55.15 % | 7.15 % | 37.70 % | 87.48 % | 0.13 % | 12.38 % | 973.8274 * | 0.4492 |
| Blacks | 54.44 % | 7.91 % | 37.65 % | 98.13 % | 0.21 % | 1.66 % | 247.2364 * | 1.1418 |
| Asians | 55.78 % | 4.02 % | 40.20 % | 88.46 % | 0.21 % | 11.32 % | 120.0751 * | 0.3700 |
| Native Americans | 64.50 % | 2.96 % | 32.54 % | 91.11 % | 0.00 % | 8.89 % | 37.1934 * | 0.2546 |
| Hispanics | 55.10 % | 4.08 % | 40.82 % | 93.64 % | 0.00 % | 6.36 % | 41.9104 * | 0.5367 |
| 18–30 years old | 58.07 % | 5.81 % | 36.12 % | 91.35 % | 0.00 % | 8.65 % | 237.1925* | 0.4798 |
| 31–45 years old | 59.18 % | 4.09 % | 36.73 % | 91.32 % | 0.00 % | 8.68 % | 338.1067 * | 0.4344 |
| 46–60 years old | 55.63 % | 7.20 % | 37.17 % | 89.12 % | 0.08 % | 10.80 % | 345.5958 * | 0.5192 |
| Over 60 | 52.94 % | 8.65 % | 38.41 % | 91.27 % | 0.30 % | 8.44 % | 616.1919 * | 0.5853 |
| Liberals | 81.79 % | 1.98 % | 16.23 % | 99.88 % | 0.00 % | 0.12 % | 316.4918 * | 0.6979 |
| Moderates | 55.03 % | 5.08 % | 39.89 % | 92.34 % | 0.09 % | 7.57 % | 414.9698 * | 0.5855 |
| Conservatives | 38.14 % | 11.66 % | 50.20 % | 70.01 % | 0.46 % | 29.53 % | 379.9408 * | 0.4119 |
| Democrats | 73.49 % | 3.03 % | 23.49 % | 99.92 % | 0.00 % | 0.08 % | 751.1915 * | 1.2404 |
| Independent | 45.05 % | 6.71 % | 48.23 % | 87.18 % | 0.32 % | 12.50 % | 243.0157 * | 0.5581 |
| Republicans | 38.34 % | 11.04 % | 50.63 % | 71.62 % | 0.45 % | 27.93 % | 521.5712 * | 0.4153 |
| Attend church | 51.44 % | 7.81 % | 40.75 % | 89.76 % | 0.12 % | 10.12 % | 860.2572 * | 0.6053 |
| Does not attend church | 59.61 % | 5.82 % | 34.57 % | 91.70 % | 0.07 % | 8.23 % | 763.7180 * | 0.4946 |
| Discusses politics | 57.33 % | 6.37 % | 36.30 % | 90.32 % | 0.24 % | 9.44 % | 1215.2496* | 0.4374 |
| Does not discuss politics | 38.27 % | 10.71 % | 51.02 % | 66.05 % | 0.53 % | 33.42 % | 77.9315 * | 0.3298 |
| Somewhat interested in politics | 59.59 % | 6.80 % | 33.61 % | 92.50 % | 0.12 % | 7.38 % | 1012.0366* | 0.5215 |
| Not very interested in politics | 45.47 % | 6.64 % | 47.89 % | 79.30 % | 0.50 % | 20.20 % | 317.4482 * | 0.3327 |

Table A16: Results of the stratified experiment on multiple-choice questions about **race diversity**. Columns "1, 2, and 3" represent the proportions of answer choices in the sample. The chi-squared test p-values <0.05 with * indicate that there is a statistically significant difference between the response distributions of the sample and ANES.

| Variable | ANES | | | Silicon | | | Chi-squared value | KL-divergence |
|---|---|---|---|---|---|---|---|---|
| | 1 | 2 | 3 | 1 | 2 | 3 | | |
| Men | 54.85 % | 13.04 % | 32.11 % | 71.16 % | 12.53 % | 16.31 % | 166.0509* | 0.0799 |
| Women | 56.85 % | 11.61 % | 31.55 % | 78.45 % | 8.10 % | 13.45 % | 309.8833 * | 0.1276 |
| Whites | 53.54 % | 12.60 % | 33.86 % | 70.08 % | 13.34 % | 16.58 % | 298.4882 * | 0.0905 |
| Blacks | 71.81 % | 11.81 % | 16.39 % | 92.37 % | 0.62 % | 7.01 % | 77.2091 * | 0.3065 |
| Asians | 57.68 % | 12.34 % | 29.97 % | 77.87 % | 3.83 % | 18.30 % | 45.3752 * | 0.1193 |
| Native Americans | 60.12 % | 8.33 % | 31.55 % | 84.69 % | 4.08 % | 11.22 % | 28.2872 * | 0.1795 |
| Hispanics | 46.46 % | 15.15 % | 38.38 % | 68.75 % | 9.82 % | 21.43 % | 10.8298 * | 0.1074 |
| 18–30 years old | 62.32 % | 12.89 % | 24.79 % | 76.85 % | 5.21 % | 17.94 % | 45.5193 * | 0.0663 |
| 31–45 years old | 54.34 % | 14.89 % | 30.78 % | 77.57 % | 7.34 % | 15.08 % | 142.1241* | 0.1312 |
| 46–60 years old | 55.68 % | 13.53 % | 30.79 % | 74.20 % | 9.08 % | 16.72 % | 91.0346 * | 0.0821 |
| Over 60 | 54.99 % | 9.23 % | 35.78 % | 71.60 % | 15.65 % | 12.75 % | 250.5528* | 0.1754 |
| Liberals | 83.38 % | 3.42 % | 13.20 % | 98.78 % | 0.31 % | 0.92 % | 236.0296 * | 0.2936 |
| Moderates | 56.89 % | 9.62 % | 33.49 % | 69.96 % | 3.29 % | 26.75 % | 59.1889 * | 0.0608 |
| Conservatives | 27.17 % | 22.53 % | 50.30 % | 41.24 % | 37.16 % | 21.60 % | 287.1184* | 0.1991 |
| Democrats | 78.61 % | 4.98 % | 16.42 % | 99.21 % | 0.16 % | 0.63 % | 535.5897 * | 0.5225 |
| Independent | 53.20 % | 10.85 % | 35.94 % | 51.36 % | 3.03 % | 45.61 % | 33.9088 * | 0.0716 |
| Republicans | 30.44 % | 21.08 % | 48.48 % | 42.97 % | 33.35 % | 23.68 % | 272.7832* | 0.1457 |
| Attend church | 48.96 % | 15.40 % | 35.64 % | 70.27 % | 11.72 % | 18.02 % | 239.4404 * | 0.1083 |
| Does not attend church | 62.18 % | 9.47 % | 28.36 % | 80.72 % | 7.13 % | 12.15 % | 244.5471 * | 0.1048 |
| Discusses politics | 56.60 % | 11.94 % | 31.46 % | 76.83 % | 9.86 % | 13.31 % | 457.4935 * | 0.1204 |
| Does not discuss politics | 48.21 % | 15.64 % | 36.15 % | 48.21 % | 9.18 % | 42.60 % | 8.6357 * | 0.0239 |
| Somewhat interested in politics | 58.64 % | 11.67 % | 29.69 % | 78.50 % | 10.72 % | 10.78 % | 400.7117* | 0.1397 |
| Not very interested in politics | 48.66 % | 13.86 % | 37.48 % | 57.49 % | 10.39 % | 32.13 % | 20.6280 * | 0.0166 |

Table A17: Results of the stratified experiment on multiple-choice questions about **health insurance**. Columns "1, 2, and 3" represent the proportions of answer choices in the sample. The chi-squared test p-values <0.05 with * indicate that there is a statistically significant difference between the response distributions of the sample and ANES.